%% file: tcs98.tex
\documentstyle[epsf]{llncs}

\include{cmds}
\include{headerlncs}

\begin{document}

\pagestyle{plain}
\author{Krzysztof R. Apt}
\institute{CWI\\
P.O. Box 94079, 1009 AB Amsterdam, The Netherlands \\
and \\
Dept. of Mathematics, Computer Science, Physics \& Astronomy \\
University of Amsterdam, The Netherlands
}
\title{The Essence of Constraint Propagation}
\maketitle
\III

\begin{abstract}
  We show that several constraint propagation algorithms (also called
  (local) consistency, consistency enforcing, Waltz, filtering or
  narrowing algorithms) are instances of algorithms that deal with
  chaotic iteration. To this end we propose a simple abstract
  framework that allows us to classify and compare these algorithms
  and to establish in a uniform way their basic properties.
\II

\NI {\em Note.} This is a full, revised version of our article ``From Chaotic
Iteration to Constraint Propagation'', Proc. of 24th International
Colloquium on Automata, Languages and Programming (ICALP '97),
(invited lecture), Springer-Verlag Lecture Notes in Computer Science
1256, pp. 36-55, (1997).
\II

\NI {\em Keywords:} constraint propagation, chaotic iteration, 
generic algorithms.
\end{abstract}

\vspace{-4mm}

\section{Introduction}

\subsection{Motivation}
Over the last ten years constraint programming emerged as an interesting
and viable approach to programming. In this approach the programming process
is limited to a generation of requirements (``constraints'') and
a solution of these requirements by means of general and domain specific 
methods.
The techniques useful for finding solutions to sets of
constraints were studied for
some twenty years in the field of Constraint Satisfaction.
One of the most important of them is
{\em constraint propagation\/}, a process of reducing
a constraint satisfaction problem to another one that is equivalent
but ``simpler''.

The algorithms that achieve such a reduction
usually aim at reaching some ``local consistency'', which 
denotes some property approximating in some loose sense
``global consistency'', which is
the consistency of the whole constraint satisfaction problem.
In fact, most of the notions of local consistency are neither
implied by nor imply global consistency (for a simple 
illustration of this  statement see, e.g., Example
\ref{exa:arccon} in Subsection \ref{subsec:auto}).

For  some constraint satisfaction problems such an enforcement
of local consistency is already sufficient for finding a solution 
in an efficient way or for
determining that none exists. 
In some other cases this process substantially reduces
the size of the search space which makes it possible to solve the original
problem more efficiently by means of some search algorithm.

The aim of this paper is to show that the constraint propagation
algorithms (also called (local) consistency, consistency enforcing,
Waltz, filtering or narrowing algorithms) can be naturally explained
by means of {\em chaotic iteration}, a basic technique used for
computing limits of iterations of finite sets of functions that
originated from numerical analysis (see, e.g., 
Chazan and Miranker \cite{CM69}) and
was adapted for computer science needs by Cousot and Cousot \cite{CC77a}.

In our presentation we study chaotic iteration of monotonic and
inflationary functions on partial orders first. This is done in
Section \ref{sec:chaotic}.  Then, in Section \ref{sec:cons} we show
how specific constraint propagation algorithms can be obtained by
choosing specific functions and specific partial orders.

This two-step presentation reveals that several constraint propagation
algorithms proposed in the literature are instances of generic chaotic
iteration algorithms studied here.

The adopted framework allows us to prove properties of these
algorithms in a simple, uniform way.  This clarifies which properties
of the so-called reduction functions (also called relaxation rules or
narrowing functions) account for correctness of these algorithms.  For
example, it turns out that idempotence is not needed here.  Further,
this framework allows us to separate an analysis of general properties, such as
termination and independence of the scheduling strategy, from
consideration of specific, constraint-related properties, such as
equivalence.  Even the consequences of choosing a queue instead of a
set for scheduling purposes can be already clarified without
introducing constraints.

We also explain how
by characterizing a given notion of a local consistency as a
common fixed point of a finite set of monotonic and inflationary
functions we can automatically generate an algorithm achieving this
notion of consistency by ``feeding'' these functions into a generic
chaotic iteration algorithm.
By studying these functions in separation we can also compare
specific constraint propagation algorithms.

A recent work of Monfroy and {R{\'{e}}ty} 
\cite{MR99} also shows how this approach
makes it possible to derive generic distributed constraint propagation
algorithms in a uniform way.

Several general presentations of constraint propagation algorithms
have been published before. In Section \ref{sec:concluding}
we explain how our work relates to and generalizes the work of others.

\subsection{Preliminaries}
\label{subsec:prel}

\begin{definition} Consider a sequence of domains ${\cal D} := D_1, \LL, D_n$.
  \begin{itemize}

  \item By a {\em scheme\/} (on $n$) we mean a sequence of different elements from $[1..n]$.
  \item 
We say that $C$ is a {\em constraint  (on ${\cal D}$) with scheme\/} 
$i_1, \LL, i_l$ if
$C \sse D_{i_1}  \times \cdots \times D_{i_l}$.

\item Let ${\bf s} := s_1, \LL, s_k$ be a sequence of schemes.
We say that a sequence of constraints  $C_1, \LL, C_k$ on ${\cal D}$ is an
{\bf s}-{\em sequence\/} if each $C_i$ is with scheme $s_i$.

\item
By a {\em Constraint Satisfaction Problem\/} $\langle \cal D; \cal C\rangle$, in short CSP, we mean
a sequence of domains ${\cal D}$ together with an {\bf s}-sequence of
constraints ${\cal C}$ on ${\cal D}$. We call then {\bf s} the {\em scheme\/} of  
$\langle \cal D; \cal C\rangle$.
\HB  
\end{itemize}
\end{definition}

In principle a constraint can have more than one scheme, for example
when all domains are equal.
This eventuality should not cause any problems in the sequel.
  Given an $n$-tuple $d := d_1, \LL, d_n$
in $D_1 \times \cdots \times D_n$ and a scheme $s := i_1, \LL, i_l$ on
$n$ we denote by $d[s]$ the tuple $d_{i_1}, \LL , d_{i_l}$.  In
particular, for $j \in [1..n]$ \ $d[j]$ is the $j$-th element of $d$.
By a {\em solution\/} to a CSP $\langle \cal D; \cal C\rangle$, where
${\cal D} := D_1, \LL, D_n$, we mean an $n$-tuple $d \in D_1 \times
\cdots \times D_n$ such that for each constraint $C$ in ${\cal C}$ with
scheme $s$ we have $d[s] \in C$.

Consider now a sequence of schemes $s_1, \LL, s_k$.  By its {\em
  union}, written as $\lan s_1, \LL, s_k \ran$ 
 we mean the scheme obtained from the sequences $s_1, \LL, s_k$ by removing from
each $s_i$ the elements present in some $s_j$, where $j < i$, and by concatenating
the resulting sequences. For example, $\lan (3,7,2), (4,3,7,5), (3,5,8)  \ran =  (3,7,2,4,5,8)$.
Recall that for an $s_1, \LL, s_k$-sequence of
constraints $C_1, \LL, C_k$
their {\em join\/}, written as  $C_1 \Join \cdots \Join C_k$,
is defined as the constraint with scheme
$\lan s_1, \LL, s_k \ran$ and such that
\[
d \in C_1 \Join \cdots \Join C_k \mbox{ iff $d[s_i] \in C_i$ for $i \in [1..k]$}.
\]

Further, given a constraint $C$ and a subsequence $s$ of its scheme,
we denote by $\Pi_{s}(C)$ the constraint with scheme $s$
defined by 
\[
\Pi_{s}(C) := \C{d[s] \mid d \in C},
\]
and call it {\em the projection of $C$ on $s$}. 
In particular, for a constraint $C$ with scheme $s$ and an element $j$ of $s$,
$\Pi_{j}(C) = \C{a \mid \te d \in C \: a = d[j]}$.

Given a CSP $\langle \cal D; \cal C\rangle$ we denote by $Sol(\langle \cal D; \cal C\rangle)$
the set of all solutions to it. If the domains are clear from the context we drop the
reference to $\cal D$ and just write $Sol({\cal C})$.
The following observation is useful.
\begin{note} \label{not:sol}
Consider a CSP $\langle \cal D; \cal C\rangle$ with 
 ${\cal D} := D_1, \LL, D_n$ and ${\cal C} := C_1, \LL, C_k$ and with
scheme {\bf s}.
\begin{enumerate} \smallromani
\item  \mbox{}\\[-9mm]
\[
Sol(\langle {\cal D}; {\cal C}\rangle) = C_1 \Join \cdots \Join C_k \Join_{i \in I} D_i,
\]
where $I := \C{i \in [1..n] \mid \mbox{ $i$ {\rm does not appear in {\bf s}}}}$.
\item
For every {\bf s}-subsequence {\bf C} of ${\cal C}$ and
$d \in Sol(\lan {\cal D}; {\cal C} \ran)$ we have
$d[\lan {\bf s} \ran] \in Sol({\bf C})$.

\HB
\end{enumerate}
\end{note}

Finally, we call two CSP's {\em equivalent\/} if they have the same
set of solutions. Note that we do not insist that these CSP's have
the same sequence of domains or the same scheme.

\section{Chaotic Iterations}
\label{sec:chaotic}

In our study of constraint propagation we proceed in two stages.  In
this section we study chaotic iterations of functions on partial
orders. Then in the next section we explain how this framework can be
readily used to explain constraint propagation algorithms.

\subsection{Chaotic Iterations on Simple Domains}
\label{subsec:ci-sd}

In general, chaotic iterations are defined for functions that
are projections on individual components 
of a specific function with several arguments.
In our approach we study a more elementary situation in which the
functions are unrelated but satisfy certain properties.
We need the following concepts.

\begin{definition}
Consider a set $D$, an element $d \in D$ and a set of functions 
 $F := \C{f_1, \LL , f_k}$ on $D$.
 \begin{itemize}
 \item 
By a {\em run\/} (of the functions $f_1, \LL, f_k$) we mean
an infinite sequence of numbers from $[1..k]$.

\item A run $i_1, i_2, \LL$ is called {\em fair\/} if 
every $i \in [1..k]$ appears in it infinitely often.

\item 
By an {\em iteration of $F$ associated with a run
$i_1, i_2, \LL$ and starting with $d$\/} 
we mean
an infinite sequence of values 
$d_0, d_1, \LL  $ defined inductively by
\[
d_0 := d,
\]
\[
d_{j} := f_{i_{j}}(d_{j-1}).
\]

When 
$d$ is the least element of $D$ in some partial
order clear from the context, we drop the reference to $d$
and talk about an {\em iteration of $F$}.

\item An iteration of $F$ is 
called {\em chaotic\/} if it is associated with 
a fair run.
\HB
 \end{itemize}
\end{definition}

\begin{definition}
Consider a partial order $(D, \po)$. A function $f$ on $D$ is called
\begin{itemize}
\item {\em inflationary\/} if 
$x \po f(x)$ for all $x$,

\item {\em monotonic\/} \index{function!monotonic}
if $x \po y$ implies 
$f(x) \po f(y)$ for all $x, y$,

\item {\em idempotent} if
$f(f(x)) = f(x)$ for all $x$.
\HB
\end{itemize}
\end{definition}

In what follows we study chaotic iterations
on specific partial orders.

\begin{definition}
We call a partial order $(D, \po )$  an {\em $\sqcup$-po\/} if
\begin{itemize}
\item $D$ contains the
least element, denoted by $\bot$, 

\item for every increasing sequence
\[
d_0 \: \po \: d_1 \: \po \: d_2 \: \LL
\]
of elements from $D$, the least upper bound of the set
\[
\C{ d_0 , \: d_1 , \: d_2 , \LL },
\]
denoted by $\bigsqcup_{n=0}^{\infty} d_n$ and called
the {\em limit of\/} $d_0, d_1, \LL$, exists,

\item for all $a,b \in D$ the least upper bound
of the set $\C{a,b}$, denoted by $a \sqcup b$, exists.
\end{itemize}

Further, we say that 
\begin{itemize}

\item an increasing sequence
$d_0 \: \po \: d_1 \: \po \: d_2 \: \LL$
{\em eventually stabilizes at d\/} if for some $j \geq 0$ we have
$d_i = d$ for $i \geq j$,

\item 
a partial order satisfies
the {\em finite chain property} if 
every increasing sequence of its elements eventually stabilizes.
\HB
\end{itemize}

\end{definition}

Intuitively, $\bot$ is an element with the least amount of information
and $a \sqsubseteq b$ means that $b$ contains more information than
$a$. Clearly, the second condition of the definition of $\sqcup$-po
is automatically satisfied if $D$ is finite.

It is also clear that $\sqcup$-po's are closed under the Cartesian product.
In the applications we shall use specific $\sqcup$-po's built out of sets
and their Cartesian products.

\begin{definition}
Let $D$ be a set.
We say that a family ${\cal F}(D)$ of subsets of $D$ is
{\em based on $D$\/} if

\begin{itemize}

\item $D \in {\cal F}(D)$,

\item for every decreasing sequence
\[
X_0 \supseteq X_1 \supseteq X_2 \LL
\]
of elements of ${\cal F}(D)$ 
\[
\cap^{\infty}_{i=0} X_i \in {\cal F}(D),
\]

\item for all $X, Y \in {\cal F}(D)$ we have $X \cap Y \in {\cal F}(D)$.
\end{itemize}

That is, a set ${\cal F}(D)$ of subsets of $D$ is based on $D$ iff
${\cal F}(D)$ with the relation $\po$ defined by
\[
X \sqsubseteq Y \mbox{ iff } X \supseteq Y
\]
is an $\sqcup$-po. 
In this $\sqcup$-po $\bot = D$ and $X \sqcup Y = X \cap Y$.
We call $({\cal F}(D), \sqsubseteq)$ an $\sqcup$-po {\em based on $D$}. 
\HB
\end{definition}

The following two examples of families of subsets based on a domain 
will be used in the sequel.

\begin{example}

Define
\[
{\cal F}(D) := {\cal P}(D),
\] 
that is ${\cal F}(D)$ consists of all subsets of $D$.
This family of subsets will be used to discuss general
constraint propagation algorithms.
\HB
\end{example}

\begin{example}
\label{exa:partial}
Let $(D, \po)$ be a partial order with 
the $\po$-least element {\em min}, the $\po$-greatest element {\em max}
and such that for every two elements $a, b \in D$
both $a \sqcup b$ and $a \sqcap b$ exists.

Examples of such partial orders are
a linear order with the $\po$-least element and the $\po$-greatest element
and the set of all subsets of a given set with the subset relation.

Given two elements $a,b$ of $D$ define
\[
[a,b] := \C{c \mid a \leq c \mbox{ and } c \leq b}
\]
and call such a set an {\em interval}.
So for $b < a$ we have $[a,b] = \ES$, for $b = a$ we have $[a,b] = \C{a}$
and $[{\em min} .. {\em max}] = D$.

Let now $F$ be a finite subset of $D$ containing {\em min} and {\em max}.
Define
\[
{\cal F}(D) := \C{[a,b] \mid a,b \in F},
\] 
that is ${\cal F}(D)$ consists of all intervals with the bounds in $F$.
Note that ${\cal F}(D)$ is indeed a
family of subsets based on $D$ since

\begin{itemize}

\item $D = [{\em min} ..{\em max}]$,

\item ${\cal F}(D)$ is finite, so every decreasing sequence of elements of ${\cal F}(D)$ 
eventually stabilizes,

\item for $a,b,c,d \in F$ we have
\[
[a,b] \cap [c,d] = [a \sqcup c, b \sqcap d].
\]
\end{itemize}

Such families of subsets will be used to
discuss constraint propagation algorithms on reals.
In these applications $D$ will be the set of real numbers augmented
with $- \infty$ and $+ \infty$ and $F$ the set of floating point numbers. 
\HB
\end{example}

The following observation can be easily distilled from a more general
result due to Cousot and Cousot \cite{CC77a}. 
To keep the paper self-contained we
provide a direct proof.

\begin{theorem}[(Chaotic Iteration)] \label{thm:chaotic}
Consider  an $\sqcup$-po $(D , \po )$
and a set of functions 
 $F := \C{f_1, \LL , f_k}$ on $D$.
Suppose that all functions in  $F$ are inflationary and monotonic.
Then the limit of every chaotic iteration of $F$ exists and
coincides with 
\[
\bigsqcup_{j=0}^{\infty} f\uparrow j,
\]
where the function $f$ on $D$ is defined by:
\[
f(x) := \bigsqcup_{i=1}^{k} f_i(x)
\]
and  $f\uparrow j$ is an abbreviation for $f^{j}(\bot)$, the
$j$-th fold iteration of $f$ started at $\bot$.
\end{theorem}
\Proof
First, notice that $f$ is inflationary, so 
$\bigsqcup_{j=0}^{\infty} f\uparrow j$ exists.
Fix a chaotic iteration $d_0, d_1, \LL$ of $F$ 
associated with a fair run $i_1, i_2, \LL$.
Since all functions $f_i$ are inflationary, 
$\bigsqcup_{j=0}^{\infty} d_j$ exists.
The result follows directly from the following two claims.

\begin{claim}
$\fa j \: \te m \: f \uparrow j \po d_m$.
\end{claim}
{\em Proof.}
We proceed by induction on $j$.
\II

\NI
{\bf Base}. $j = 0$.
As $f \uparrow 0 = \bot = d_0$, the claim is obvious.
\II

\NI
{\bf Induction step}.
Assume that for some $j \geq 0$ we have
$f \uparrow j \po d_m$ for some $m \geq 0$.
Since

\[
f \uparrow (j+1) = f(f \uparrow j) = \bigsqcup_{i=1}^{k} f_i(f \uparrow j), 
\]
it suffices to prove 
\begin{equation}
\fa i \in [1..k] \: \te m_i \: f_i(f \uparrow j) \po d_{m_i}.
\label{equ:incl}
\end{equation}
Indeed, we have then by the fact that
$d_l \po d_{l+1}$ for $l \geq 0$ 
\[
\bigsqcup_{i=1}^{k} f_i(f \uparrow j) \po \bigsqcup_{i=1}^{k} d_{m_i} \po d_{m'}
\]
where $m' := max \C{m_i \mid i \in [1..k]}$.

So fix $i \in [1..k]$. By fairness of the considered run
 $i_1, i_2, \LL$,
for some $m_{i} > m$
we have $i_{m_{i}} = i$. 
Then $d_{m_i} = f_i(d_{m_{i} -1})$.
Now  $d_m \po d_{m_{i} -1}$, so
by the monotonicity of $f_i$ we have

\[
f_i(f \uparrow j) \po f_i(d_m) \po f_i(d_{m_{i} -1}) = d_{m_i}.
\]
This proves (\ref{equ:incl}).
\HB
\III

\begin{claim}
$\fa m  \: d_m \po f \uparrow m$.
\end{claim}
{\em Proof.}
The proof is by a straightforward induction on $m$.
Indeed, for $m = 0$ we have
$d_0  = \bot =  f \uparrow 0$, so the induction base holds.

To prove the induction step 
suppose that for some $m \geq 0$
we have $d_m \po f \uparrow m$. For some $i \in [1..k]$ 
we have $d_{m+1} = f_i(d_m)$, so by the monotonicity of $f$ we get
$
d_{m+1}  = f_i(d_m) \po f(d_m)\po f(f \uparrow m) = f \uparrow (m+1).
$
\HB

\HB
\VV

In many situations some chaotic iteration studied in the Chaotic
Iteration Theorem \ref{thm:chaotic} eventually stabilizes. 
This is for example the case when
$(D, \po )$ satisfies the finite chain property.
In such cases the limit of every chaotic iteration
can be characterized in an alternative way.

\begin{corollary}[(Stabilization)] \label{cor:chaotic}
Suppose that 
under the assumptions of the Chaotic Iteration Theorem \ref{thm:chaotic}
some chaotic iteration of $F$ eventually stabilizes. Then
every chaotic iteration of $F$ eventually stabilizes at
the least fixed point of $f$.
\end{corollary}
\Proof
It suffices to note that if some chaotic iteration 
$d_0, d_1 \LL$ of $F$ eventually stabilizes at some $d_m$
then by Claims 1 and 2 $f \uparrow m = d_m$, so

\begin{equation}
\bigsqcup_{j=0}^{\infty} f\uparrow j = f \uparrow m.
\label{eq:stab}
\end{equation}
Then, again by Claims 1 and 2, every chaotic iteration 
of $F$ stabilizes at  $f \uparrow m$
and it is easy to see that by virtue of (\ref{eq:stab})
$f \uparrow m$ is the least fixed point of $f$.
\HB
\VV

Finally, using the above results we can compare chaotic iterations
resulting from different sets of functions.

\begin{corollary}[(Comparison)] \label{cor:chaotic2}
Consider  an $\sqcup$-po $(D , \po )$
and two set of functions, 
$F := \C{f_1, \LL , f_k}$ and  $G := \C{g_1, \LL , g_l}$ on $D$.
Suppose that all functions in $F$ and $G$ are inflationary and monotonic.
Further, assume that for $i \in [1..k]$ there exist $j_1, \LL, j_m \in [1..l]$
such that
\[
f_i(x) \po g_{j_1} \circ \LL \circ g_{j_m}(x) \mbox{ for all $x$.}
\]
Then $lim(F) \po lim(G)$ for the uniquely defined
limits $lim(F)$ and $lim(G)$ of the 
chaotic iterations of $F$ and $G$.
\end{corollary}
\Proof
Straightforward using the Chaotic
Iteration Theorem \ref{thm:chaotic}
and the fact that the functions in $G$ are inflationary.
\HB  

\subsection{Chaotic Iterations on Compound Domains} 
\label{subsec:ci-cd}
Not much more can be deduced about the process of the chaotic iteration 
unless the structure of the domain $D$ is further known.
So assume now that $\sqcup$-po $(D, \po )$ is the  
Cartesian product of the $\sqcup$-po's
$(D_i, \po_i )$, for $i \in [1..n]$.
In what follows we consider a modification of the situation studied in 
the Chaotic Iteration Theorem \ref{thm:chaotic} in which each
function $f_i$ affects only certain components of $D$.

Consider the partial orders
$(D_i, \po_i )$, for $i \in [1..n]$ and
a scheme $s := i_1, \LL, i_l$ on $n$.
Then by $(D_s, \po_s)$ we mean the Cartesian product of the
partial orders
$(D_{i_j}, \po_{i_j})$, for $j \in [1..l]$.

Given a function $f$ on $D_s$ we say that $f$ is {\em with scheme $s$}. 
Instead of defining iterations for the case of the functions with schemes,
we rather reduce the situation to the one studied in the previous subsection.
To this end we canonically extend each function $f$ on $D_s$
to a function $f^+$ on $D$ as follows. 
Suppose that $s = i_1,  \LL, i_l$ and
\[
f(d_{i_1}, \LL, d_{i_l}) = (e'_{i_1}, \LL, e'_{i_l}).
\]
Let for $j \in [1..n]$
\[
e_j := \left \{ \begin{array}{ll}
                   e'_j     & \mbox{if $j$ is an element of $s$}, \\
                   d_j          & \mbox{otherwise. }
                                      \end{array}
                            \right.
\]
Then we set
\[
f^+(d_1, \LL,  d_n) := (e_1, \LL,   e_n).
\]

Suppose now that  $(D, \po )$ is the Cartesian product of the
$\sqcup$-po's $(D_i, \po_i )$, for $i \in [1..n]$, and $F := \C{f_1, \LL, f_k}$ 
is a set of functions with schemes that are all inflationary and monotonic.
Then the following algorithm can be used
to compute the limit of the chaotic iterations of $F^+ := \C{f^+_1, \LL,f^+_k}$.
We say here that a function $f$
{\em depends on $i$\/} if $i$ is an element of its scheme.
\II

\NI
{\sc Generic Chaotic Iteration Algorithm ({\tt CI})}
\begin{tabbing}
\= $d := \underbrace{(\bot, \LL, \bot)}_{\mbox{$n$ times}}$; \\[1mm]
\> $d' := d$; \\ 
\> $G := F$; \\ 
\> {\bf while} $G \neq \ES$ {\bf do} \\
\> \qquad choose $g \in G$; suppose $g$ is with scheme $s$; \\
\> \qquad $G := G - \C{g}$; \\
\> \qquad $d'[s] := g(d[s])$; \\
\> \qquad {\bf if} $d[s] \neq d'[s]$ {\bf then} \\
\> \qquad \qquad $G := G \cup \C{f \in  F \mid \mbox{$f$ depends on
 some } i \mbox{ in $s$ such that } d[i] \neq d'[i]}$; \\
\> \qquad \qquad $d[s] := d'[s]$ \\
\> \qquad {\bf fi} \\
\> {\bf od} 
\end{tabbing}

Obviously, the condition $d[s] \neq d'[s]$ can be
omitted here. We retained it to keep the form of the
algorithm more intuitive.

The following observation will be useful in
the proof of correctness of this algorithm.

\begin{note} \label{not:extend}
Consider the partial orders
$(D_i, \po_i )$, for $i \in [1..n]$,
a scheme $s$ on $n$
and a function $f$ with scheme $s$. Then
\begin{enumerate}\smallromani
\item $f$ is inflationary iff $f^+$ is.

\item $f$ is monotonic iff $f^+$ is.
\HB
\end{enumerate}
\end{note}

Observe that, in spite of the name of the algorithm, its infinite
executions do not need to correspond to chaotic iterations. 
The following example will be of use for a number of different purposes.

\begin{example} \label{exa:inf}
Consider the set of natural numbers $\cal N$ augmented with $\omega$, with
the order $\leq$. In this order $k \leq \omega$ for $k \in {\cal N}$.
Next, we consider the following three functions on ${\cal N} \cup \C{\omega}$:

\[
f_1(n) := \left \{ \begin{array}{ll}
                   n+1     & \mbox{if $n$ is even}, \\
                   n       & \mbox{if $n$ is odd}, \\
                   \omega & \mbox{if $n$ is $\omega$, }
                                      \end{array}
                            \right.
\]

\[
f_2(n) := \left \{ \begin{array}{ll}
                   n+1     & \mbox{if $n$ is odd}, \\
                   n       & \mbox{if $n$ is even}, \\
                   \omega & \mbox{if $n$ is $\omega$, }
                                      \end{array}
                            \right.
\]
\[
f_3(n) := \omega.
\]

\NI
Clearly, the underlying order is an $\sqcup$-po and the functions
$f_1, f_2$ and $f_3$ are all inflationary, monotonic and idempotent.
Now, there is an infinite execution of the {\tt CI} algorithm
that corresponds with the run $1,2,1,2, \LL$.
This execution does not correspond to any chaotic iteration
of $\C{f_1, f_2, f_3}$.
\HB
\end{example}

However, when we focus on terminating executions we obtain 
the following result in the proof of which our analysis of chaotic
iterations is of help.

\begin{theorem}[({\tt CI})] \label{thm:CI}
  \mbox{} \\[-6mm]
\begin{enumerate}\smallromani
\item Every terminating execution of the {\tt CI} algorithm computes
in $d$ the least fixed point of
the function $f$ on $D$ defined by
\[
f(x) := \bigsqcup_{i=1}^{k} f^+_i(x).
\]

\item  If all $(D_i, \po_i )$, where $i \in [1..n]$, 
satisfy the finite chain property, then every execution of the 
{\tt CI} algorithm terminates.
  \end{enumerate}
\end{theorem}

\Proof
It is simpler to reason about a modified, but equivalent, algorithm
in which the assignments $d'[s] := g(d[s])$ and $d[s] := d'[s]$ are
respectively replaced by $d' := g^+(d)$ and $d := d'$ and the test
$d[s] \neq d'[s]$ by $d \neq d'$. 
\II

\NI
$(i)$
Note that the formula
\[
I :=\fa f \in F - G \:  f^+(d) = d
\]
is an invariant of the {\bf while} loop of the modified algorithm.
Thus upon its termination
\[
(G = \ES) \A I
\]
holds, that is  
\[
\fa f \in F \: f^+(d) = d.
\]
Consequently, some chaotic iteration of $F^+$ eventually stabilizes at $d$.
Hence $d$ is the least fixpoint of the function $f$ defined in item $(i)$
because the
Stabilization Corollary \ref{cor:chaotic} is applicable here by
virtue of Note \ref{not:extend}.
\II

\NI
$(ii)$
Consider the lexicographic order of the partial orders
$(D, \sqsupseteq)$ and $({\cal N}, \leq)$, 
defined on the elements of $D \times {\cal N}$ by
\[ 
(d_1, n_1) \leq_{lex} (d_2, n_2)\ {\rm iff} \ d_1 \sqsupset d_2
        \ {\rm or}\ ( d_1 = d_2 \ {\rm and}\ n_1 \leq n_2). 
\]
We use here the inverse order $\sqsupset$
defined by: $d_1 \sqsupset d_2$ iff $d_2 \sqsubseteq d_1$ and $d_2 \neq d_1$.

By Note \ref{not:extend}(i) all functions $f^+_i$
are inflationary, so
with each {\bf while} loop iteration of the modified algorithm
the pair
\[
(d, card \: G)
\]
strictly decreases in this order $\leq_{lex}$.
However, in general the lexicographic order $(D \times {\cal N}, \leq_{lex})$ is not
well-founded and in fact termination is not guaranteed.
But assume now additionally that each partial order
$(D_i, \po_i )$ satisfies the finite chain property. Then so does their
Cartesian product $(D, \po )$.
This means that $(D, \sqsupseteq)$ is well-founded and consequently so is
$(D \times {\cal N}, \leq_{lex})$ which implies termination.
\HB
\VV

When all considered functions $f_i$ are also idempotent,
we can reverse the order of
the two assignments to $G$, that is to put the assignment
$G := G - \C{g}$ after the {\bf if-then-fi} statement,
because after applying an idempotent function there is
no use in applying it immediately again.
Let us denote by {\tt CII} the algorithm resulting from this movement of
the assignment $G := G - \C{g}$.

More specialized versions of the {\tt CI} and {\tt CII} algorithms
can be obtained by representing $G$ as a queue. To this end we use the operation
${\bf enqueue}(F, Q)$ which for a set $F$ and a queue $Q$ enqueues in an 
arbitrary order all the elements of $F$ in $Q$,
denote the empty queue by {\bf empty}, and the head and the tail 
of a non-empty queue $Q$ respectively by ${\bf head}(Q)$ and ${\bf tail}(Q)$.
The following algorithm is then a counterpart of the {\tt CI} algorithm.
\II

\NI
{\sc Generic Chaotic Iteration Algorithm with a Queue ({\tt CIQ})}
\begin{tabbing}
\= $d := \underbrace{(\bot, \LL, \bot)}_{\mbox{$n$ times}}$; \\[1mm]
\> $d' := d$; \\ 
\> $Q := {\bf empty}$; \\
\> ${\bf enqueue}(F, Q)$; \\
\> {\bf while} $Q \neq {\bf empty}$ {\bf do} \\
\> \qquad $g := {\bf head}(Q)$; suppose $g$ is with scheme $s$; \\
\> \qquad $Q := {\bf tail}(Q)$; \\
\> \qquad $d'[s] := g(d[s])$; \\
\> \qquad {\bf if} $d[s] \neq d'[s]$ {\bf then} \\
\> \qquad \qquad ${\bf enqueue}(\C{f \in  F \mid \mbox{$f$ depends on
 some } i \mbox{ in $s$ such that } d[i] \neq d'[i]}, Q)$; \\
\> \qquad \qquad $d[s] := d'[s]$ \\
\> \qquad {\bf fi} \\
\> {\bf od} 
\end{tabbing}

Denote by {\tt CIIQ} the modification of  the {\tt CIQ} algorithm
that is appropriate for the idempotent functions, so the one in which the 
assignment $Q := {\bf tail}(Q)$ is performed after the {\bf if-then-fi} statement.

It is easy to see that the claims of the {\tt CI}
Theorem \ref{thm:CI} also hold for the
{\tt CII, CIQ} and {\tt CIIQ} algorithms.
A natural question arises whether for the specialized versions
{\tt CIQ} and {\tt CIIQ}
some additional properties can be established. The answer is positive. 
We need an auxiliary notion and a result first.

\begin{definition}
Consider a set of functions $F := \C{f_1, \LL, f_k}$ on a domain $D$.

\begin{itemize}
\item We say that an element $i \in [1..k]$ is {\em eventually 
irrelevant for an iteration $d_0, d_1, \LL$ of $F$\/} if
$\te m \geq 0 \: \fa j \geq m \: f_i(d_j) = d_j$.

\item An iteration of $F$ is called {\em semi-chaotic\/} if every
$i \in [1..k]$ that appears finitely often in its run is eventually 
irrelevant for this iteration.
\HB
\end{itemize}
\end{definition}

So every chaotic iteration is semi-chaotic but not conversely.

\begin{note} \label{note:semi}
  \mbox{} \\[-6mm]
\begin{enumerate}\smallromani
\item Every semi-chaotic iteration $\xi$ corresponds to a chaotic
  iteration $\xi'$ with the same limit as $\xi$ and such that $\xi$
  eventually stabilizes at some $d$ iff $\xi'$ does.
\item   Every infinite execution of the {\tt CIQ}  (respectively {\tt CIIQ})
algorithm corresponds to a semi-chaotic iteration.

  \end{enumerate}

\end{note}
\Proof \\
{\em (i)\/}
$\xi$ can
be transformed into the desired chaotic iteration $\xi'$
by repeating from a certain moment on some elements of it.
\II

\NI
{\em (ii)\/}
Consider an infinite execution of the {\tt CIQ} algorithm. 
Let $i_1, i_2, \LL$ be the run associated with it and
$\xi := d_0, d_1, \LL$ the iteration of $F^+$ associated
with this run.

Consider the set $A$ of the elements of $[1..k]$ that appear 
finitely often in the run  $i_1, i_2, \LL$. 
For some $m \geq 0$ we have $i_j \not\in A$ for $j > m$.
This means by the structure of this algorithm
that after $m$ iterations of the {\bf while} loop no function $f_i$
with $i \in A$ is ever present in the queue $Q$.

By virtue of the invariant $I$ used in the proof of the {\tt CI}
Theorem \ref{thm:CI} we then have 
$f^+_i(d_j) = d_j$ for $i \in A$ and $j \geq m$.
This proves that $\xi$ is semi-chaotic. 

The proof for the {\tt CIIQ} algorithm is the same.
\HB
\VV

Item {\em (i)\/} shows that the results of Subsection \ref{subsec:ci-sd}
can be strengthened to semi-chaotic iterations.  However, the 
property of being a semi-chaotic iteration cannot be determined from
the run only. So, for simplicity, we decided to limit our exposition
to chaotic iterations. Next, it is easy to
show that item {\em (ii)\/} cannot be strengthened to chaotic
iterations.

We can now prove the desired results. The first one shows that the
nondeterminism present in the {\tt CIQ} and {\tt CIIQ} algorithms has
no bearing on their termination.

\begin{theorem}[(Termination)] \label{thm:CIQ}
If some execution of the {\tt CIQ} (respectively {\tt CIIQ})
algorithm terminates, then
all executions of the {\tt CIQ} (respectively {\tt CIIQ})
algorithm terminate.
\end{theorem}
\Proof
We concentrate on the {\tt CIQ} algorithm. For the {\tt CIIQ} algorithm
the proof is the same. 

Consider a terminating execution of the {\tt CIQ} algorithm.
Construct a chaotic iteration of $F^+$ the initial prefix of which
corresponds with this execution.  By virtue of the invariant $I$ this
iteration eventually stabilizes.  By the Stabilization Corollary
\ref{cor:chaotic}
\begin{equation}
\mbox{every chaotic iteration of $F^+$ eventually stabilizes.}
\label{eq:stab1}
\end{equation}

Suppose now by contradiction that some execution of the {\tt CIQ}
algorithm does not terminate. Let $\xi$ be the iteration of $F^+$
associated with this execution.
By the structure of this algorithm
\begin{equation}
\mbox{$\xi$ does not eventually stabilize.}
\label{eq:stab2}
\end{equation}

By Note \ref{note:semi}(ii) $\xi$ is a semi-chaotic iteration.
Consider a chaotic iteration $\xi'$ of $F^+$ that corresponds with
$\xi$ by virtue of Note \ref{note:semi}(i). We conclude by
(\ref{eq:stab2}) that $\xi'$ does not eventually stabilize. This
contradicts (\ref{eq:stab1}).  \HB \VV

So for a given Cartesian product $(D, \po )$ of the  $\sqcup$-po's
and a finite set $F$ of inflationary, monotonic and idempotent 
functions either all executions
of the  {\tt CIQ} (respectively {\tt CIIQ}) algorithm terminate or
all of them are infinite. In the latter case we can be more specific.

\begin{theorem}[(Non-termination)] \label{thm:limit}
  For every infinite execution of the {\tt CIQ} (respectively {\tt
    CIIQ}) algorithm the limit of the corresponding iteration of $F$
  exists and coincides with
\[
\bigsqcup_{j=0}^{\infty} f\uparrow j,
\]
where $f$ is defined as in the  {\tt CI}
Theorem \ref{thm:CI}(i).
\end{theorem}
\Proof 
Consider an infinite execution of the {\tt CIQ} algorithm.  By
Note \ref{note:semi}(ii) it corresponds to a semi-chaotic iteration $\xi$ of
$F^+$.  
By Note \ref{note:semi}(i)  $\xi$ corresponds to a chaotic iteration
of $F^+$ with the same limit. The desired conclusion now follows by
the Chaotic Iteration Theorem \ref{thm:chaotic}.

The proof for the {\tt CIIQ} algorithm is the same.
\HB 
\VV

Neither of the above two results holds for the {\tt CI} and {\tt CII}
algorithms.  Indeed, take the $\sqcup$-po $({\cal N} \cup \C{\omega}, \leq)$
and the functions $f_1, f_2, f_3$ of Example \ref{exa:inf}.  Then
clearly both infinite and finite executions of the {\tt CI} and {\tt
  CII} algorithms exist.  We leave to the reader the task of modifying
Example \ref{exa:inf} in such a way that for both {\tt CI} and {\tt
  CII} algorithms infinite executions exist with different limits
of the corresponding iterations.

\section{Constraint Propagation}
\label{sec:cons}
Let us return now to the study of CSP's. We show here how the results
of the previous section can be used to explain the constraint
propagation process.

In general, two basic approaches fall under this name:

\begin{itemize}

\item reduce the constraints while maintaining equivalence;

\item reduce the domains while maintaining equivalence.
\end{itemize}

\subsection{Constraint Reduction} 
\label{subsec:cr}

In each step of the constraint reduction process one or more
constraints are replaced by smaller ones.  In general, the smaller
constraints are not arbitrary. For example, when studying linear
constraints usually the smaller constraints are also linear.

To model this aspect of constraint reduction 
we associate with each CSP an $\sqcup$-po
that consists of the CSP's that can be generated during the
constraint reduction process.

Because the domains are assumed to remain unchanged, we can
identify each CSP with the sequence of its constraints.
This leads us to the following notions.

Consider a CSP ${\cal P} := \langle  {\cal D}; C_1, \LL, C_k\rangle$. Let
for $i \in [1..k]$ 
$({\cal F}(C_i), \supseteq)$ be an $\sqcup$-po based on $C_i$. 
We call the Cartesian product
$(CO, \po)$ of $({\cal F}(C_i), \supseteq)$, with $i \in [1..k]$,
{\em a constraint $\sqcup$-po associated with ${\cal P}$}.

As in Subsection \ref{subsec:ci-cd}, for
a scheme $s := i_1,  \LL, i_l$ we denote by 
$(CO_s, \po_s)$ the Cartesian product of the partial orders
$({\cal F}(C_{i_j}), \supseteq)$, where $j \in [1..l]$.

Note that 
$CO_s = {\cal F}(C_{i_1}) \times \cdots \times {\cal F}(C_{i_l})$.
Because we want now to use constraints in our analysis
and constraint are sets of tuples, we identify $CO_s$ with the set
\[
\C{ X_1 \times \cdots \times X_l \mid \mbox{ $X_j \in {\cal F}(C_{i_j})$ for $j \in [1..l]$}}.
\]
In this way we can write the elements of $CO_s$ as 
Cartesian products
$X_1 \times \cdots \times X_l$, so as
(specific) sets of $l$-tuples, 
instead of as $(X_1, \LL, X_l)$,
and similarly with $CO$.

Note that $C_1 \times \cdots \times C_k$ is the $\po$-least element of $CO$.
Also, note that because of the use of the inverse subset order
$\supseteq$
we have for
$X_{1} \times \cdots \times X_{l} \in CO_s$ and
$Y_{1} \times \cdots \times Y_{l} \in CO_s$
\begin{center}
  
\begin{tabular}{lll}
$X_1 \times \cdots \times X_l \po_s Y_1 \times \cdots \times Y_l$ & \ iff &
$X_1 \times \cdots \times X_l \supseteq Y_1 \times \cdots \times Y_l$ \\
& (iff & $X_i \supseteq Y_i$ for $i \in [1..l]$),
\end{tabular}
\end{center}
\begin{center}
\begin{tabular}{lll}
$(X_1 \times \cdots \times X_l) \sqcup_s (Y_1 \times \cdots \times Y_l)$ & \ = &
$(X_1 \times \cdots \times X_l) \cap (Y_1 \times \cdots \times Y_l)$ \\
 & (= & $(X_1 \cap Y_1) \times \cdots \times (X_l \cap Y_l))$.
\end{tabular}
\end{center}

This allows us to use from now on the
set theoretic counterparts $\supseteq$ and $\cap$ of $\po_s$ and $\sqcup_s$.
Note that for the partial order $(CO_s, \po_s)$
a function $g$ on $CO_s$ is inflationary iff 
${\bf C} \supseteq g({\bf C})$ and $g$ is monotonic iff it is monotonic w.r.t.
the  set inclusion.

So far we have introduced an $\sqcup$-po associated with a CSP. Next, we
introduce functions by means of which chaotic iterations will
be generated.

\begin{definition} \label{def:crf}
  Consider a CSP $\lan {\cal D}; C_1, \LL, C_k \ran$ together with a
  sequence of families of sets ${\cal F}(C_i)$ based on $C_i$, for $i
  \in [1..k]$, and a scheme $s$ on $k$. By a {\em constraint
    reduction function with scheme $s$\/} we mean a function $g$ on
  $CO_s$ such that for all ${\bf C} \in CO_s$
\begin{itemize}
\item  ${\bf C} \supseteq g({\bf C})$,

\item  $Sol({\bf C}) = Sol(g({\bf C}))$.
\HB
\end{itemize}
\end{definition}

{\bf C} is here a Cartesian product of some constraints and 
in the second condition we
identified it with the sequence of these constraints,
and similarly with $g({\bf C})$.
The first condition states that $g$ reduces the constraints
$C_i$, where $i$ is an element of $s$,
while the second condition states that during this 
constraint reduction
process no solution to ${\bf C}$ is lost.

\begin{example} \label{exa:projection}
  As a first example of a constraint reduction function
take  ${\cal F}(C) :=  {\cal P}(C)$ for each constraint $C$
and consider the following function $g$ on some $CO_s$:

\[
g(C \times {\bf C}  ) := C' \times {\bf C},
\]
where $C' = \Pi_t(Sol(C, {\bf C}))$ and $t$ is the scheme of
$C$. 
In other words,  $C'$ is the projection of the set of solutions
of $(C, {\bf C})$ on the scheme of $C$.

To see that $g$ is indeed a constraint reduction function,
first note that
by the definition of $Sol$ we have $C' \sse C$, so
$C \times {\bf C} \supseteq g(C \times {\bf C})$.
Next, note that for $d \in Sol(C, {\bf C})$ we have
$d[t] \in \Pi_t(Sol(C, {\bf C}))$,
so $d \in Sol(C', {\bf C})$.
This implies that 
$Sol(C, {\bf C}) = Sol(g(C, {\bf C})).$

Note also that $g$ is monotonic w.r.t. the set inclusion and idempotent.
\HB
\end{example}

\begin{example} \label{exa:path}
  As another example that is of importance for the discussion in
  Subsection \ref{subsec:related} consider a CSP $\langle D_1, \LL,
  D_n; \cal C\rangle$ of binary constraints such that for each scheme
  $i,j$ on $n$ there is exactly one constraint, which we denote by
  $C_{i,j}$.  Again put ${\cal F}(C) := {\cal P}(C)$ for each
  constraint $C$.

Define now for each scheme $k,l,m$ on $n$ 
the following function $g^m_{k,l}$ on $CO_s$, where
$s$ is the triple corresponding to the positions of the constraints
$C_{k,l}, C_{k,m}$ and  $C_{m,l}$ in ${\cal C}$:

\[
g^m_{k,l}(X_{k,l} \times X_{k,m} \times X_{m,l}) := 
(X_{k,l} \cap \Pi_{k,l}(X_{k,m} \Join X_{m,l})) \times  X_{k,m} \times X_{m,l}.
\]

To prove that the functions  $g^m_{k,l}$ are
constraint reduction functions it suffices to note that
by simple properties of the $\Join$ operation and by
Note \ref{not:sol}(i) we have
\begin{center}
  
\begin{tabular}{lll}
$X_{k,l} \cap \Pi_{k,l}(X_{k,m} \Join X_{m,l})$ &  = & $\Pi_{k,l}(X_{k,l}  \Join X_{k,m} \Join X_{m,l})$ \\
                                            & =  & $\Pi_{k,l}(Sol(X_{k,l}, X_{k,m}, X_{m,l}))$,
\end{tabular}
\end{center}
so these functions are special cases of the functions defined in
Example \ref{exa:projection}.
\HB  
\end{example}

\begin{example} \label{exa:cuts}

As a final example consider linear inequalities over integers.
Let $x_1, \LL ,x_n$ be different variables ranging over integers,
where $n > 0$.
By a {\em linear inequality\/} we mean here a formula of the
form
\[
\sum_{i = 1}^{n} a_i x_i \leq b,
\]
where $a_1, \LL ,a_n$ and $b$ are integers.

In what follows we consider CSP's that consist of finite
or countable sets of linear inequalities. Each such set determines
a subset of ${\cal N}^n$ which we view as a single constraint.
Call such a subset an {\em INT-LIN\/} set.

Fix now a constraint $C$ that is an {\em INT-LIN\/} set formed by a
finite or countable set {\em LI\/} of linear inequalities.  Define
${\cal F}(C)$ to be the set of {\em INT-LIN\/} sets formed by a finite
or countable set of linear inequalities extending {\em LI}.
Clearly, ${\cal F}(C)$ is a family of sets based on $C$.

Given now $m$ linear inequalities
\[
\sum_{i = 1}^{n} a^{j}_{i}  x_{i} \leq b^{j},
\]
where $j \in [1..m]$, and $m$ nonnegative reals $c_1, \LL ,c_m$, we
construct a new linear inequality
\[
\sum_{i = 1}^{n} (\sum_{j = 1}^{m} c_j a^{j}_{i})  x_{i} \leq \sum_{j = 1}^{m} c_j b^{j}.
\]

If for $i \in [1..n]$ each coefficient
$\sum_{j = 1}^{m}  c_j a^{j}_{i}$ is an integer, then
we replace the right-hand side by  $\floor{\sum_{i = 1}^{m} c_j b^{j}}$.

This yields the inequality
\[
\sum_{i = 1}^{n} (\sum_{j = 1}^{m} c_j a^{j}_{i})  x^{j}_{i} \leq \floor{\sum_{j = 1}^{m} c_j b^{j}}
\]
that is called a {\em Gomory-Chv\'{a}tal cutting plane}.

An addition of a cutting plane to a set of linear inequalities on
integers maintains equivalence, so it is an example of a constraint
reduction function. 

It is well-known that the process of deriving cutting planes does not
have to stop after one application (see, e.g., 
Cook, Cunningham, Pulleyblank, and Schrijver \cite[Section
6.7]{CCPS98}), so this reduction function is non-idempotent.
\HB
\end{example}

We now show that when the constraint reduction function discussed in
Example \ref{exa:projection} is modified by applying it to
each argument constraint
simultaneously, it becomes a constraint reduction function that
is in some sense optimal. 

More precisely, assume
the notation of Definition \ref{def:drf} and let $s := i_1,  \LL, i_l$.
Define a function $\rho$
on $CO_s$ as follows:
\[
\rho({\bf C}) := {\bf C}',
\]

\NI
where 
\[
{\bf C}  := C_{i_1} \times \cdots \times C_{i_l},
\]

\[
{\bf C}'  := C'_{i_1} \times \cdots \times C'_{i_l},
\]
with each $C'_{i_j} :=  \Pi_{t_j}(Sol({\bf C}))$,
where $t_j$ is the scheme of  $C_{i_j}$.

So $\rho({\bf C})$ replaces every constraint $C$ in {\bf C} 
by  the projection of $Sol({\bf C})$ on the scheme of $C$.

\begin{note}[(Characterization)] \label{note:crf-char}
Assume the notation of Definition \ref{def:crf}.
A function $g$ on $CO_s$ is a constraint reduction function 
iff for all ${\bf C} \in  CO_s$
\[
\rho({\bf C}) \sse g({\bf C}) \sse {\bf C}.
\]
\end{note}
\Proof
Suppose that $s := i_1,  \LL, i_l$.
We have the following string of equivalences for 
\[
g({\bf C}) := X_{i_1} \times \cdots \times X_{i_l}:
\]
$\rho({\bf C}) \sse g({\bf C})$ iff 
$\Pi_{t_j}(Sol({\bf C})) \sse X_{i_j}$ for $j \in [1..l]$ iff
$Sol({\bf C}) \sse Sol(g({\bf C}))$.

So 
$\rho({\bf C}) \sse g({\bf C}) \sse {\bf C}$ iff
($Sol({\bf C}) = Sol(g({\bf C}))$ and $g({\bf C}) \sse {\bf C}$).
\HB
\VV

Take now a  CSP ${\cal P} := \langle  {\cal D}; C_1, \LL, C_k\rangle$
and a sequence of constraints $C'_1, \LL, C'_k$ such that $C'_i \sse C_i$ for
$i \in [1..k]$.
Let ${\cal P}' := \langle  {\cal D}; C'_1, \LL, C'_k\rangle$.
We say then that ${\cal P'}$ {\em is determined by ${\cal P}$
and $C'_1 \times \cdots \times C'_k$}.
Further, we say that ${\cal P'}$ is {\em smaller than\/} ${\cal P'}$ and
${\cal P}$ is {\em larger than\/} ${\cal P'}$.

Consider now a  CSP ${\cal P} := \langle  {\cal D}; C_1, \LL, C_k\rangle$
and a constraint reduction function $g$.
Suppose that 
\[
g^+(C_1 \times \cdots \times C_k) = C'_1 \times \cdots \times C'_k,
\]
where $g^+$ is the canonic extension of $g$ to $CO$ defined 
in Subsection \ref{subsec:ci-cd}.
We now define 
\[
g({\cal P}) := \langle  {\cal D}; C'_1, \LL, C'_k\rangle.
\]
We have the following observation.

\begin{lemma} \label{lem:c-equ}
Consider a CSP ${\cal P}$ and a constraint reduction function $g$.
Then ${\cal P}$ and $g({\cal P})$ are equivalent.
\end{lemma}
\Proof
Suppose that $s$ is the scheme of the function $g$ and let
{\bf C} be an element of $CO_s$. So {\bf C} is a Cartesian product
of some constraints.
As before we identify it with the sequence of these constraints. 
For some sequence of schemes
{\bf s}, \ {\bf C} is
the {\bf s}-sequence of the
constraints of ${\cal P}$.

Let now $d$ be a solution to ${\cal P}$.
Then by Note \ref{not:sol}(ii) we have 
$d[\lan {\bf s} \ran] \in Sol({\bf C})$, so by the definition of $g$ also
$d[\lan {\bf s} \ran] \in Sol(g({\bf C}))$.
Hence for every constraint $C'$ in $g({\bf C})$ with scheme $s'$ we have 
$d[s'] \in C'$ since
$d[\lan {\bf s} \ran][s'] = d[s']$. So
$d$ is a solution to $g({\cal P})$.
The converse implication holds by the definition of a constraint reduction function.
\HB
\VV

When dealing with a specific CSP with a constraint $\sqcup$-po
associated with it we have in general several constraint reduction
functions, each defined on a possibly different domain.  To study the
effect of their interaction we can use the Chaotic Iteration Theorem
\ref{thm:chaotic} in conjunction with the above Lemma.  After
translating the relevant notions into set theoretic terms we get the
following direct consequence of these results.  (In this translation
$CO_s$ corresponds to $D_s$ and $CO$ to $D$.)

\begin{theorem}[(Constraint Reduction)] \label{thm:cons}
Consider a CSP ${\cal P} := \langle  {\cal D}; C_1, \LL, C_k\rangle$
with a constraint $\sqcup$-po associated with it.
Let $F := \C{g_1, \LL , g_k}$, where each $g_i$ is a 
constraint reduction function. 
Suppose that all functions $g_i$ are monotonic
w.r.t. the set inclusion. Then 
\begin{itemize}
\item the limit of 
every chaotic iteration of 
$F^+ := \C{g^+_1, \LL , g^+_k}$
exists;
\item this limit coincides with 
\[
\bigcap_{j=0}^{\infty} g^{j}(C_1 \times \cdots \times C_k),
\]
where the function $g$ on $CO$ is defined by:
\[
g({\bf C}) := \bigcap_{i=1}^{k} g^+_i({\bf C}),
\]

\item the CSP determined by ${\cal P}$ and this limit is equivalent to ${\cal P}$.
\HB
\end{itemize}
\end{theorem}

Informally, this theorem states that the order of the applications of the
constraint reduction functions does not matter, as long as none of them is
indefinitely neglected.
Moreover, the CSP corresponding to the limit of such an iteration
process of the constraint reduction functions is equivalent to the
original one.

Consider now a CSP ${\cal P}$ with a constraint $\sqcup$-po associated with it
that satisfies the finite chain property. Then we can use the
{\tt CI, CII, CIQ} and {\tt CIIQ} algorithms to compute the limits of
the chaotic iterations considered in the above Theorem.
We shall explain in Subsection \ref{subsec:related} how
by instantiating these algorithms with specific constraint $\sqcup$-po's and
constraint reduction functions
we obtain specific algorithms considered in the literature.

In each case, by virtue of the {\tt CI} Theorem \ref{thm:CI} and its reformulations
for the {\tt CII, CIQ} and {\tt CIIQ} algorithms, we can conclude that
these algorithms compute the greatest common
fixpoint w.r.t. the set inclusion of the functions from $F^+$.
Consequently, the CSP determined by ${\cal P}$ and this limit is the 
largest CSP that is both smaller than ${\cal P}$ and is a fixpoint of
the considered constraint reduction functions.

So the limit of the constraint propagation process
could be added to the collection of important greatest fixpoints 
presented in Barwise and Moss \cite{BM96}.

\subsection{Domain Reduction}
\label{subsec:dr}

In this subsection we study the domain reduction
process. First, we associate with each CSP an $\sqcup$-po
that ``focuses'' on the domain reduction.

Consider a CSP ${\cal P} := \langle  D_1, \LL, D_n;  {\cal C} \rangle$. Let
for $i \in [1..n]$ 
$({\cal F}(D_i), \supseteq)$ be an $\sqcup$-po based on $D_i$. 
We call the Cartesian product
$(DO, \po)$ of $({\cal F}(D_i), \supseteq)$, with $i \in [1..n]$
{\em a domain $\sqcup$-po associated with ${\cal P}$}.

As in Subsection \ref{subsec:ci-cd}, for
a scheme $s := i_1,  \LL, i_l$ we denote by 
$(DO_s, \po_s)$ the Cartesian product of the partial orders
$({\cal F}(D_{i_j}), \supseteq)$, where $j \in [1..l]$.
Then, as in the previous subsection, we identify $DO_s$ with the set
\[
\C{ X_1 \times \cdots \times X_l \mid \mbox{ $X_j \in {\cal F}(D_{i_j})$ for $j \in [1..l]$}}.
\]

Next, we introduce functions that reduce domains.
These functions are associated with constraints. Constraints are arbitrary
sets of $k$-tuples for some $k$, while the $\po_s$ order and the $\sqcup_s$ 
operation are defined only on Cartesian
products. So to define these functions we use the
set theoretic counterparts $\supseteq$ and $\cap$ of $\po_s$ and $\sqcup_s$
which are defined on arbitrary sets.

\begin{definition} \label{def:drf}
Consider a sequence of domains $D_1, \LL, D_n$
together with a sequence of families of sets  ${\cal F}(D_i)$ based on $D_i$, 
for  $i \in [1..n]$, and a scheme $s$ on $n$.
By a {\em domain reduction function\/} for a constraint $C$ with scheme $s$
we mean a function $f$ on $DO_s$ such that for all ${\bf D} \in DO_s$
\begin{itemize}
\item  ${\bf D} \supseteq f({\bf D})$,

\item  $C \cap {\bf D} = C \cap f({\bf D})$.
\HB
\end{itemize}
\end{definition}

The first condition states that $f$ reduces the ``current'' domains
associated with the constraint $C$
(so no solution to $C$ is ``gained''),
while the second condition states that during this 
domain reduction process no solution to $C$ is ``lost''.
In particular, the second condition implies that 
if $C \sse {\bf D}$ then
$C \sse f({\bf D})$.

\begin{example} \label{exa:arc}
As a simple example of a domain reduction functions consider a binary constraint
$C \sse D_1 \times D_2$. 
Let ${\cal F}(D_i) := {\cal P}(D_i)$ with $i \in [1,2]$ 
be the families of sets based on $D_1$ and $D_2$.
 
Define now the projection functions $\pi_1$ and $\pi_2$ on
$DO_{1,2} = {\cal P}(D_1) \times {\cal P}(D_2)$ as follows:
\[
\pi_1(X \times Y) := X' \times Y,
\]
where $X' = \C{a \in X \mid \te b \in Y \: (a,b) \in C}$, and
\[
\pi_2(X \times Y) := X \times Y',
\]
where $Y' = \C{b \in Y \mid \te a \in X \: (a,b) \in C}$.
It is straightforward to check that $\pi_1$ and $\pi_2$ are
indeed domain reduction functions.
Further, these functions are
monotonic w.r.t. the  set inclusion and idempotent.
\HB
\end{example}

\begin{example}  \label{exa:arcn}
As another example of a domain reduction function 
consider an $n$-ary constraint
$C \sse D_1 \times \cdots \times D_n$.
Let for $i \in [1..n]$ the family of sets based on $D_i$
be defined by
${\cal F}(D_i) := {\cal P}(D_i)$.

Note that 
$DO = {\cal P}(D_1) \times \cdots \times {\cal P}(D_n)$.
Define now the projection function $\pi_{C}$ by putting for ${\bf D} \in DO$

\[
\pi_{C}({\bf D}) := \Pi_1(C \cap {\bf D}) \times \cdots \times \Pi_n(C \cap {\bf D}).
\]
Recall from  Subsection \ref{subsec:prel}
that  $\Pi_i(C \cap {\bf D}) = \C{a \mid \te d \in C \cap {\bf D} \ a = d[i]}$.
Clearly $\pi_{C}$ is a domain reduction function for $C$ and is
monotonic w.r.t. the  set inclusion and idempotent.

Here the scheme of $C$ is $1,\LL, n$. Obviously, $\pi_{C}$
can be defined in an analogous way for a constraint $C$ with an 
arbitrary scheme.
\HB  
\end{example}

\begin{figure}[htbp]
  \begin{center}
    \leavevmode
\epsfxsize7cm
\centerline{\epsfbox{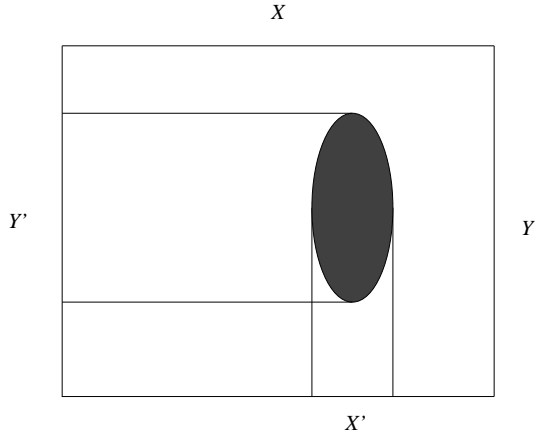}}
    \caption{Domain reduction functions.}
    \label{fig:projection}
  \end{center}
\end{figure}

So all three domain reduction functions deal with projections, respectively
on the first, second or all components
 and
can be visualized by means of Figure
\ref{fig:projection}.

The following observation provides an equivalent definition of a domain
reduction function in terms of the projection function defined in the
last example. 

\begin{note}[(Characterization)] \label{note:char}
Assume the notation of Definition \ref{def:drf}.
A  function $f$ on $DO_s$ is a domain reduction function 
for the constraint $C$ iff
for all ${\bf D} \in DO_s$
\[
\pi_{C}({\bf D}) \sse f({\bf D}) \sse {\bf D}.
\]
\end{note}
\Proof
Suppose that $s := i_1,  \LL, i_l$.
We have the following string of equivalences for 
\[
f({\bf D}) := X_{i_1} \times \cdots \times X_{i_l}:
\]
$\pi_{C}({\bf D}) \sse f({\bf D})$ iff
$\Pi_{i_j}(C \cap {\bf D}) \sse X_{i_j}$ for $j \in [1..l]$ iff
$C \cap {\bf D} \sse f({\bf D})$.

So $\pi_{C}({\bf D}) \sse f({\bf D}) \sse {\bf D}$ iff
($C \cap {\bf D} = C \cap f({\bf D})$ and $f({\bf D}) \sse {\bf D}$).
\HB
\VV

Intuitively, this observation means that the projection function
$\pi_{C}$ is an ``optimal'' domain reduction function.  In general,
however, $\pi_{C}$ does not need to be a domain reduction function, since
the sets $\Pi_i(C \cap {\bf D})$ do not have to belong to the
used families of sets based on the domain $D_i$.  The next
example provides an illustration of such a situation.

\begin{example}
\label{exa:reals}
Consider an $n$-ary constraint $C$ on reals, that is
$C \sse {\cal R}_{+}^{n}$.
Let ${\cal R}_{+} := {\cal R} \cup \C{+ \infty, - \infty}$,
$F$ be a finite subset of ${\cal R}_{+}$ containing 
$- \infty$ and $+ \infty$
and let the family 
${\cal F}({\cal R}_{+})$ of subsets of ${\cal R}_{+}$
be defined as in Example \ref{exa:partial}.
So 
\[
{\cal F}({\cal R}_{+}) = \C{[a,b] \mid a,b \in F}
\]
and
\[
DO = \C{[a_1, b_1] \times \cdots \times [a_n, b_n] \mid a_i,b_i \in F \mbox{ for } i \in [1..n]}.
\]

Further, given a subset $X$ of ${\cal R}_{+}$ we define
\[
int(X) := \cap \C{Y \in {\cal F}({\cal R}_{+}) \mid X \subseteq Y}.
\]
So $int(X)$ is the smallest interval with bounds in $F$ that contains $X$.
Clearly, $int(X)$ exists for every $X$.

Define now the function $f$ on $DO$ by putting for ${\bf D} \in DO$

\[
f({\bf D}) := int(\Pi_1(C \cap {\bf D})) \times \cdots \times int(\Pi_n(C \cap {\bf D})).
\]

Benhamou and Older \cite{BO97} proved that $f$ is a domain reduction function that is
monotonic w.r.t. the set inclusion and idempotent.
Note that the first property is a direct consequence of the
Characterization Note \ref{note:char}.
\HB
\end{example}

All the domain reduction functions given so far
were idempotent. We now provide an example of a natural non-idempotent
reduction function. 

\begin{example}
\label{exa:lineq}
We consider linear equalities  over integer interval domains.
By a {\em linear equality\/} we mean here a formula of the
form
\[
\sum_{i = 1}^{n} a_i x_i = b,
\]
where $a_1, \LL ,a_n$ and $b$ are integers.

In turn, by an {\em integer interval\/} we mean
an expression of the form
\[
[a..b]
\]
where $a$ and $b$ are integers;
$[a..b]$ denotes the set of all integers between $a$ and $b$,
including $a$ and $b$.

The domain reduction functions for linear equalities over
integer intervals are simple modifications of the reduction rule introduced in
Davis \cite[page 306]{davis87} that dealt with linear constraints over
closed intervals of reals.
In the case of a linear equality
\[
\sum_{i \in {\em POS}} a_i x_i -  \sum_{i \in {\em NEG}} a_i x_i = b
\]
where 
\begin{itemize}
\item $a_i$ is a positive integer for $i \in {\em POS} \cup {\em NEG}$,

\item $x_i$ and $x_j$ are different variables for $i \neq j$ and $i,j
  \in {\em POS} \cup {\em NEG}$,

\item $b$ is an integer,
\end{itemize}
such a function is defined as follows (see, e.g., Apt \cite{Apt98a}):

\[
f([l_1 .. h_1],  \LL,  [l_n..h_n]) := 
([l'_1 .. h'_1],  \LL,  [l'_n..h'_n])
\]
where for $j \in {\em POS}$
\[
l'_j := max(l_j, \ceiling{\gamma_j}), \ h'_j :=  min(h_j, \floor{\alpha_j}),
\]
for $j \in {\em NEG}$
\[
l'_j := max(l_j, \ceiling{\beta_j}),  \ h'_j := min(h_j, \floor{\delta_j}),
\]
and where
\[
\alpha_j := \frac{b - \sum_{i \in {\em POS} - \{j\}} a_i l_i +  \sum_{i \in {\em NEG}} a_i h_i}{a_j}
\]

\[
\beta_j := \frac{- b + \sum_{i \in {\em POS}} a_i l_i -  \sum_{i \in {\em NEG} - \{j\}} a_i h_i}{a_j}
\]

\[
\gamma_j := \frac{b - \sum_{i \in {\em POS} - \{j\}} a_i h_i +  \sum_{i \in {\em NEG}} a_i l_i}{a_j}
\]
and 
\[
\delta_j := \frac{- b + \sum_{i \in {\em POS}} a_i h_i -  \sum_{i \in {\em NEG} - \{j\}} a_i l_i}{a_j}
\]
(It is worthwhile to mention that this function can be derived by
means of cutting planes mentioned in Example \ref{exa:cuts}).

Fix now some initial integer intervals $I_1, \LL, I_n$  and
let for $i \in [1..n]$ the family of sets ${\cal F}(I_i)$ consist
of all integer subintervals of $I_i$.

The above defined function $f$ is then a domain reduction function
defined on  the Cartesian product of ${\cal F}(I_i)$ for $i \in [1..n]$
and is easily seen to be non-idempotent.  For example, in case of the
CSP
\[
{\p{x \in [0..9], y \in [1..8]}{3x - 5y = 4}}
\]
a straightforward calculation shows that 
\[
f([0..9], [1..8]) = ([3..9], [1..4])
\]
and
\[
f([3..9], [1..4]) = ([3..8], [1..4]).
\]
\HB  
\end{example}

Take now a  CSP ${\cal P} := \langle D_1, \LL, D_n; {\cal C}\rangle$ 
and a sequence of domains $D'_1, \LL, D'_n$ such that $D'_i \sse D_i$ for
$i \in [1..n]$.
Consider a CSP ${\cal P'}$ obtained from ${\cal P}$ by replacing
each domain $D_i$ by $D'_i$ and by restricting each constraint in ${\cal C}$ to
these new domains. We say then that ${\cal P'}$ {\em is determined by ${\cal P}$
and $D'_1 \times \cdots \times D'_n$}.

Consider now a  CSP ${\cal P} := \langle D_1, \LL, D_n; {\cal C}\rangle$ 
with a domain $\sqcup$-po associated with it
and a domain reduction function $f$ for a constraint $C$ of ${\cal C}$.
We now define $f({\cal P})$ to be the CSP obtained from ${\cal P}$ 
by reducing its domains using the function $f$. 

More precisely, suppose that 
\[
f^+(D_1 \times \cdots \times D_n) = D'_1 \times \cdots \times D'_n,
\]
where $f^+$ is the canonic extension of $f$ to $DO$ defined 
in Subsection \ref{subsec:ci-cd}.
Then $f({\cal P})$ is the CSP determined by ${\cal P}$ 
and $D'_1 \times \cdots \times D'_n$. The following observation is
an analogue of Lemma \ref{lem:c-equ}.
\begin{lemma}
Consider a CSP ${\cal P}$ and a domain reduction function $f$.
Then ${\cal P}$ and $f({\cal P})$ are equivalent.
\end{lemma}

\Proof
Suppose that $D_1, \LL, D_n$ are
the domains of ${\cal P}$ and assume that $f$ is a domain reduction function 
for $C$ with scheme $i_1, \LL, i_l$. 
By definition $f$ is defined on $D_{i_1} \times \cdots \times D_{i_l}$.
Let
\[
f(D_{i_1} \times \cdots \times D_{i_l}) = D'_{i_1} \times \cdots \times D'_{i_l}.
\]
Take now a solution $d$ to ${\cal P}$.  Then 
$d[i_1, \LL, i_l] \in C$, so by the definition of $f$ also 
$d[i_1, \LL, i_l] \in D'_{i_1} \times \cdots \times D'_{i_l}$. So $d$ is
also a solution to $f({\cal P})$.
The converse implication holds by the definition of a domain reduction function.
\HB
\VV

Finally, the following result is an analogue of the Constraint
Reduction Theorem \ref{thm:cons}.  It is a consequence of Iteration
Theorem \ref{thm:chaotic} and the above Lemma, obtained by translating
the relevant notions into set theoretic terms.
(In this translation $DO_s$ corresponds to $D_s$ and $DO$ to $D$.)

\begin{theorem}[(Domain Reduction)] \label{thm:dom}
Consider a CSP ${\cal P} := \langle D_1, \LL, D_n; \cal C\rangle$
with a domain $\sqcup$-po associated with it.
Let $F := \C{f_1, \LL , f_k}$, where each $f_i$ is a 
domain reduction function for some constraint in
${\cal C}$.  Suppose that all functions $f_i$ are monotonic
w.r.t. the set inclusion. Then 
\begin{itemize}
\item the limit of 
every chaotic iteration of 
$F^+ := \C{f^+_1, \LL , f^+_k}$
exists;
\item this limit coincides with 
\[
\bigcap_{j=0}^{\infty} f^{j}(D_1 \times \cdots \times D_n),
\]
where the function $f$ on $DO$ is defined by:
\[
f({\bf D}) := \bigcap_{i=1}^{k} f^+_i({\bf D}),
\]

\item the CSP determined by ${\cal P}$ and this limit is equivalent to ${\cal P}$.
\HB
\end{itemize}
\end{theorem}

The above result shows an analogy between the domain reduction
functions. In fact, the domain reduction functions can be modeled as
constraint reduction functions in the following way.

First, given a CSP $\langle D_1, \LL, D_n; {\cal C} \rangle$
add to it $n$ unary constraints, each of which coincides with a
different domain $D_i$. This yields
${\cal P} := \langle D_1, \LL, D_n; {\cal C},  D_1, \LL, D_n \rangle$.
Obviously, both CSP's are equivalent.

Next, associate, as in the previous
subsection, with each constraint $C$ of ${\cal P}$
an $\sqcup$-po ${\cal F}(C)$ based on it. 

Take now a constraint $C \in {\cal C}$ with a 
scheme $s := i_1,  \LL, i_l$ and a function $f$ on $DO_s$.
Define a function $g$ on 

\[
{\cal F}(C) \times {\cal F}(D_{i_1}) \cdots \times {\cal F}(D_{i_l})
\]
by
\[
g(C', {\bf D}) := (C', f({\bf D})).
\]

Then $f$ is a domain reduction function iff
$g$ is a constraint reduction function, since
$Sol(C', {\bf D}) := C' \cap {\bf D}$.

This simple representation of the domain reduction functions as the
constraint reduction functions shows that the latter concept is more
general and explains the analogy between the results on the constraint
reduction functions and domain reduction functions.
It also allows us to analyze the outcome of ``hybrid'' chaotic
iterations in which both domain reduction functions and constraint
reduction functions are used.

We discussed the domain reduction functions separately,
because, as we shall see in the next section, they have
been extensively studied, especially in the context of CSP's with
binary constraints and of interval arithmetic.

\subsection{Automatic Derivation of Constraint Propagation Algorithms}
\label{subsec:auto}

We now show how specific provably correct algorithms for achieving a
local consistency notion can be automatically derived.  The idea is
that we characterize a given local consistency notion as a
common fixpoint of a finite set of monotonic, inflationary
and possibly idempotent
functions and then instantiate any of the 
{\tt CI, CII, CIQ} or {\tt CIIQ} algorithms with these functions.
 As it is difficult to define local consistency formally,
we illustrate the idea on two examples.

\begin{example}
\label{exa:arccon}
First, consider the notion of arc-consistency for $n$-ary relations,
defined in Mohr and Masini \cite{MM88}.
We say that a constraint $C \sse D_1 \times \cdots \times D_n$ is
{\em arc-consistent\/} if for every $i \in [1..n]$ and $a \in D_i$
there exists $d \in C$ such that $a = d[i]$.
That is, for every involved domain each element of it participates in a
solution to $C$.
A CSP is called {\em arc consistent\/} if every constraint 
of it is.

For instance, the CSP $\lan \C{0,1}, \C{0,1}; =, \neq \ran$ that
consists of two binary constraints, that of equality and inequality
over the 0-1 domain, is arc consistent (though obviously
inconsistent).

Note that a CSP $\lan D_1, \LL, D_n ; {\cal C}\ran$ 
is arc consistent iff for every constraint $C$ of it
with a scheme $s := i_1,  \LL, i_l$ we have
$\pi_C(D_{i_1}  \times \cdots \times D_{i_l}) = D_{i_1}  \times \cdots \times D_{i_l}$,
where $\pi_C$ is defined in Example \ref{exa:arcn}.
We noted there that the projection functions
$\pi_C$ are domain reduction functions that are monotonic w.r.t. the
set inclusion and idempotent.

By virtue of the {\tt CI} Theorem \ref{thm:CI} reformulated for the {\tt CII} algorithm,
we can now use the {\tt CII} algorithm to achieve arc consistency 
for a CSP with finite domains by instantiating the
functions of this algorithm with the projection functions $\pi_C$.

By the Domain Reduction Theorem
\ref{thm:dom} we conclude that the CSP computed by this
algorithm is equivalent to the original one
and is the greatest arc consistent CSP that is smaller than
the original one.
\HB
\end{example}

\begin{example}
Next, consider the notion of relational consistency proposed
in Dechter and van Beek \cite{DvB97}.
Relational consistency is
a very powerful concept that generalizes several
consistency notions discussed until now.

To define it we need to introduce some auxiliary concepts first.
Consider a CSP $\lan D_1, \LL, D_n ; {\cal C}\ran$.
Take a scheme $t := i_1, \LL, i_l$ on $n$.
We call 
$d \in D_{i_1} \times \cdots \times D_{i_l}$ a tuple of {\em type $t$}.
Further, we say that $d$ is
{\em consistent\/} if for every 
subsequence $s$ of $t$ and a constraint $C \in {\cal C}$ 
with scheme $s$ we have $d[s] \in C$.

A CSP ${\cal P}$ is called {\em relationally $m$-consistent} if for
any {\bf s}-sequence $C_1, \LL, C_m$ of different constraints of
${\cal P}$ and a subsequence $t$ of $\lan {\bf s} \ran$, every
consistent tuple of type $t$ belongs to $\Pi_t(C_1 \Join \cdots \Join
C_m)$, that is, every consistent tuple of type $t$ can be extended to
an element of $Sol(C_1, \LL, C_m)$.

As the first step we characterize this notion 
as a common fixed point
of a finite set of monotonic and inflationary functions.

Consider a CSP ${\cal P} := \lan D_1, \LL, D_n; C_1, \LL, C_k \ran$.
Assume for simplicity that for every scheme $s$ on $n$ there is a
unique constraint with scheme $s$. Each CSP is trivially equivalent
with such a CSP --- it suffices to replace for each scheme $s$ the set
of constraints with scheme $s$ by their intersection and to introduce
``universal constraints'' for the schemes without a constraint.
By a ``universal constraint'' we mean here a Cartesian product of some 
domains.

Consider now a scheme $i_1, \LL, i_m$ on $k$.
Let {\bf s} be such that $C_{i_1}, \LL, C_{i_m}$ is an
{\bf s}-sequence of constraints and let 
$t$ be a subsequence of $\lan {\bf s} \ran$.
Further, let $C_{i_0}$ be the constraint of $\cal P$ with scheme $t$.
Put
$s := \lan (i_0), (i_1, \LL, i_m) \ran$.
(Note that if $i_0$ does not appear in $i_1, \LL, i_m$ then
$s = i_0, i_1, \LL, i_m$ and otherwise $s$ is the permutation of 
$i_1, \LL, i_m$ obtained by transposing $i_0$ with the first element.)

Define now a function $g_{s}$ on $CO_{s}$ by
\[
g_{s}(C \times {\bf C}) := (C \cap \Pi_t(\Join {\bf C})) \times {\bf C}.
\]
It is easy to see that 
if for each function $g_s$ of the above form we have
\[
g^+_s(C_1 \times \cdots \times C_k) = C_1 \times \cdots \times C_k,
\]
then ${\cal P}$ is relationally $m$-consistent.
(The converse implication is in general not true).
Note that the functions $g_s$ are inflationary and monotonic
w.r.t. the inverse subset order $\supseteq$ and also idempotent. 

Consequently, again by the {\tt CI} Theorem \ref{thm:CI}
reformulated for the {\tt CII} algorithm,
we can use the {\tt CII} algorithm to achieve relational $m$-consistency 
for a CSP with finite domains
by ``feeding'' into this algorithm the above defined functions.
The obtained algorithm improves upon the (authors' terminology)
brute force algorithm proposed in Dechter and van Beek 
\cite{DvB97} since the useless constraint modifications 
are avoided.

As in Example \ref{exa:path}, by simple properties of the $\Join$
operation and by Note \ref{not:sol}(i) we have 
\[
C \cap \Pi_t(\Join {\bf C}) = \Pi_t(C \Join (\Join {\bf C})) = 
\Pi_t(Sol(C, {\bf C})).
\]
Hence, by virtue of Example
\ref{exa:projection}, the functions $g_s$ are all constraint reduction
functions.  Consequently, by the Constraint Reduction Theorem
\ref{thm:cons} we conclude that the CSP computed by the just
discussed algorithm is equivalent to the original one.  
\HB
\end{example}

\section{Concluding Remarks}
\label{sec:concluding}

\subsection{Related Work}
\label{subsec:related}

As already mentioned in the introduction, the idea of chaotic
iterations was originally used in numerical analysis.  The concept
goes back to the fifties and was successively generalized into the
framework of  Baudet \cite{Bau78} on which Cousot and Cousot
\cite{CC77a} was based.
Our notion of chaotic iterations on partial orders is derived from the
last reference.
A historical overview can be found in Cousot \cite{Cou78}.

Let us turn now to a review of the work on constraint propagation.
We show how our results provide a uniform framework
to explain and generalize the work of others.

It is illuminating to see how the attempts of finding general principles
behind the constraint propagation algorithms repeatedly reoccur
in the literature on constraint satisfaction problems spanning 
the last twenty years.

As already stated in the introduction, the aim of the constraint
propagation algorithms is most often to achieve some form of local
consistency.  As a result these algorithms are usually called in the
literature ``consistency algorithms'' or ``consistency enforcing
algorithms'' though, as already mentioned, some
other names are also used.

In an early work of Montanari \cite{montanari-networks} the notion
of path-consistency was defined and a constraint propagation 
algorithm was introduced to achieve it.
Then, in the context of analysis of polyhedral scenes,
another constraint propagation algorithm was proposed in 
Waltz \cite{waltz75}. 

In Mackworth \cite{mackworth-consistency} the notion of
arc-consistency was introduced and Waltz' algorithm was explained in
more general terms of CSP's with binary constrains. Also, a unified
framework was proposed to explain the arc- and path-consis\-tency
algorithms. To this end the arc-consistency algorithm {\tt AC-3} and
the path-consistency algorithm {\tt PC-2} were introduced and the
latter algorithm was obtained from the former one by pursuing the
analogy between both notions of consistency.

A version of {\tt AC-3} consistency algorithm can be obtained by instantiating
the {\tt CII} algorithm with the domain reduction functions defined
in Example \ref{exa:arc}, whereas a version of {\tt PC-2} algorithm can be
obtained by instantiating this algorithm with the constraint reduction
functions defined in Example \ref{exa:path}.

In Davis \cite{davis87} another generalization of Waltz algorithm was
proposed that dealt with $n$-ary constraints.  The algorithm proposed
there can be obtained by instantiating the {\tt CIQ} algorithm with
the projection functions of Example \ref{exa:arc} generalized to $n$-ary
constraints. To obtain a precise match the {\bf enqueue} operation in this
algorithm should enqueue the projection functions related to one constraint
in ``blocks''.

In Dechter and Pearl \cite{dechter88} the notions of arc- and path-consistency
were modified to directional arc- and path-consistency, versions
that take into account some total order $<_d$ of the domain indices,
and the algorithms for achieving these forms of consistency were
presented.  Such algorithms can be obtained as instances of the {\tt
  CIQ} algorithm as follows.

For the case of directional arc-consistency the queue in this
algorithm should be instantiated with the set of the domain reduction
functions $\pi_1$ of Example \ref{exa:arc} for the constraints the
scheme of which is consistent with the $<_d$ order.  These functions
should be ordered in such a way that the domain reduction functions
for the constraint with the $<_d$-large second index appear earlier.
This order has the effect that the first argument
of the {\bf enqueue} operation within
the {\bf if-then-fi} statement always consists of domain
reduction functions that {\em are already\/} in the queue.
So this   {\bf if-then-fi} statement can be deleted.
Consequently, the algorithm can be rewritten as a simple {\bf for}
loop that processes the selected domain reduction functions $\pi_1$ in the
appropriate order.

For the case of directional path-consistency the constraint reduction
functions $g^m_{k,l}$ should be used only with $k,l <_d m$ and the
queue in the {\tt CIQ} algorithm should be initialized in such a way
that the functions $g^m_{k,l}$ with the $<_d$-large $m$ index appear
earlier.  As in the case of directional arc-consistency this algorithm
can be rewritten as a simple {\bf for} loop.

In Montanari and Rossi \cite{MR91} a general study of constraint propagation was
undertaken by defining the notion of a relaxation rule and by
proposing a general relaxation algorithm.  The notion of a relaxation
rule coincides with our notion of a constraint propagation function
instantiated with the functions defined in Example
\ref{exa:projection} and the general relaxation algorithm is the
corresponding instance of our {\tt CI} algorithm.

In Montanari and Rossi \cite{MR91} it was also shown that the notions
of arc-consistency and path-consistency can be defined by means of
relaxation rules and that as a result arc-consistency and
path-consistency algorithms can be obtained by instantiating with
these rules their general relaxation algorithm.

Another, early attempt at providing a general framework to explain
constraint propagation was
undertaken in Caseau \cite{caseau91}. In this paper abstract
interpretations  and a version of the {\tt CIQ} algorithm are used to
study iterations that result from applying approximations
of the projection functions of Example \ref{exa:arc} generalized 
to $n$-ary constraints.
It seems that for finite domains these approximation functions coincide with 
our concept of domain reduction functions.

Next, Van Hentenryck, Deville and Teng
\cite{vanhentenryck-generic} presented a generic arc consistency
algorithm, called {\tt AC-5}, that can be specialized to the known
arc-consistency algorithms {\tt AC-3} and {\tt AC-4} and also to new
arc-consistency algorithms for specific classes of constraints.
More recently, this work was extended in 
Deville, Barette and Van Hentenryck 
\cite{DBV97} to path-consistency algorithms.

Let us turn now our attention to constraints over reals.
In Lhomme \cite{Lho93} the notion of arc B-consistency was introduced and
an algorithm proposed that enforces it for constraint satisfaction problems
defined on reals. This algorithm can be obtained by
instantating our {\tt CI} algorithm with the functions
defined in Example \ref{exa:reals}.

Next, in Benhamou, McAllester, and Van Hentenryck 
\cite{BMV94} and Benhamou and Older \cite{BO97} specific
functions, called narrowing functions, were 
associated with constraints in the context of interval arithmetic
for reals and some properties of them were
established. In our terminology it
means that these are idempotent and monotonic domain reduction functions.
One of such functions is defined in Example \ref{exa:arcn}.
As a consequence, the algorithms proposed
in these papers, called respectively a fixpoint algorithm
and a narrowing algorithm, become the
instances of our {\tt CIIQ} algorithm and  {\tt CII} algorithm.

Other two attempts to provide a general setting for
constraint propagation algorithms can be found in
Benhamou \cite{Ben96} and Telerman and Ushakov \cite{TU96}. In these papers 
instead of $\sqcup$-po's specific
families of subsets of the considered domain are taken with the
inverse subset order. In Benhamou \cite{Ben96} they are called approximate
domains and in Telerman and Ushakov \cite{TU96} subdefinite models. 
Then specific algorithms are used to compute the outcome of
constraint propagation.
The considered families of subsets correspond to our $\sqcup$-po's, 
the discussed functions are in our terminology 
idempotent and monotonic domain restriction
functions and the considered algorithms are respectively the
instances of our {\tt CII} and {\tt CI} algorithm.

In both papers it was noted that the algorithms compute the same value
independently of the order of the applications of the functions used.
In Benhamou \cite{Ben96} local consistency is defined as the largest
fixpoint of such a collection of functions and it is observed that on finite
domains the  {\tt CII} algorithm computes this largest fixpoint.
In Telerman and Ushakov 
\cite{TU96} the subdefinite models are discussed as a general
approach to model simulation, imprecise data and constraint programming. Also
related articles that were published in 80s  in Russian are there discussed.

The importance of fairness for the study of constraint propagation was
first noticed in  G{\"u}sgen and Hertzberg \cite{guesgen-fundamental}
where chaotic iterations of monotonic domain reduction functions were considered.
Results of Section \ref{sec:chaotic} (in view of their applications to the
domain reduction process in Subsection \ref{subsec:dr}) 
generalize the results of this paper to arbitrary 
$\sqcup$-po's and their Cartesian products. In particular, 
Stabilization Corollary \ref{cor:chaotic} generalizes the main result of this paper.

Fairness also plays a prominent role 
in  Montanari and Rossi \cite{MR91}, while the relevance of the chaotic
iteration was independently noticed in Fages, Fowler, and Sola \cite{FFS96} and
van Emden \cite{Emd97}.  In the latter paper the generic chaotic iteration
algorithm {\tt CII} was formulated and proved correct for the domain
reduction functions defined in Benhamou and Older \cite{BO97} and it was
shown that the limit of the constraint propagation process for these
functions is their greatest common fixpoint.

The idea that the meaning of a constraint is a function (on a
constraint store) with some algebraic properties was put forward in
Saraswat, Rinard, and Panangaden \cite{saraswat-semantic}, where the
properties of being inflationary (called there extensive), monotonic
and idempotent were singled out.

A number of other constraint propagation algorithms that were proposed
in the literature, for example, in four out the first five issues of
the Constraints journal, can be shown to be instances of the generic
chaotic iteration algorithms.

In each of the discussed algorithms a minor optimization can be incorporated
the purpose of which is to stop the computation as soon as one of the
variable domains becomes empty. In some of the algorithms discussed
above this optimization is already present. For simplicity we disregarded
it in our discussion.
This modification can be easily incorporated into our generic algorithms
by using $\po$-po's with the greatest element $\top$ and by
enforcing an exit from the {\bf while} loop as soon as one of the 
components of $d$ becomes $\top$.

\subsection{Idempotence}

In most of the above papers the (often implicitly) considered semantic,
constraint or domain reduction functions are idempotent, so we now
comment on the relevance of this assumption.

To start with, we exhibited in Example \ref{exa:cuts} and
\ref{exa:lineq} natural constraint and domain reduction functions
that are not idempotent.  Secondly, as noticed in
Older and Vellino \cite{older-constraint}, another paper on constraints for
interval arithmetic on reals, we can always replace each
non-idempotent inflationary function $f$ by
\[
f^{*}(x) := \bigsqcup_{i=1}^{\infty} f^{i}(x).
\]
The following is now straightforward to check.

\begin{note}
Consider an $\sqcup$-po $(D, \po)$ and a function $f$ on $D$.
\begin{itemize}
\item If $f$ is inflationary, then so is $f^{*}$. 

\item If $f$ is monotonic, then so $f^{*}$. 

\item If $f$ is inflationary and $(D, \po)$ has the finite chain property, 
then $f^{*}$ is idempotent.

\item If $f$ is idempotent, then $f^{*} = f$.

\item Suppose that $(D, \po)$ has the finite chain property.
Let $F := \C{f_1, \LL, f_k}$ be a set of inflationary, monotonic
functions on $D$ 
and let $F^{*} := \C{f^{*}_1, \LL, f^{*}_k}$.
Then the limits of all
chaotic iterations of $F$ and of $F^{*}$ exist and always coincide.
\HB
\end{itemize}
\end{note}

Consequently, under the conditions of the last item,
every chaotic iteration of $F^{*}$ can be modeled by 
a chaotic iteration of $F$, though 
not conversely. In fact, the use of $F^{*}$ instead of $F$ can lead to a
more limited number of chaotic iterations. This may mean that in 
some specific algorithms
some more efficient chaotic iterations of $F$ cannot be
realized when using $F^{*}$.
For specific functions, for instance those 
studied in  Examples \ref{exa:cuts} and \ref{exa:lineq},
the computation by means of  $F^{*}$ instead of $F$
imposes a forced delay on the application of other reduction functions.

\subsection{Comparing Constraint Propagation Algorithms}

The {\tt CI} Theorem \ref{thm:CI} and its
reformulations for the {\tt CII, CIQ} and {\tt CIIQ} algorithms allow
us to establish equivalence between these algorithms. More precisely,
these result show that in case of termination all four algorithms
compute in the variable $d$ the same value.

In specific situations it is natural to consider various domain
reduction or constraint reduction functions.  When the adopted
propagation algorithms are instances of the generic algorithms here
studied, we can use the Comparison Corollary \ref{cor:chaotic2} to
compare their outcomes. By way of example
consider two instances of the {\tt CII} algorithm: one in which for
some binary constraints the pair of the domain reduction functions
defined in Example \ref{exa:arc} is used, and another in which for
these binary constraints the domain reduction function defined in
Example \ref{exa:arcn} is used.

We now prove that in case of termination both algorithms compute in
$d$ the same value. Fix a binary constraint $C$ and adopt the notation
of Example \ref{exa:arc} and of Example \ref{exa:arcn} used with
$n=2$.  Note that for ${\bf X} \in DO_{1,2}$
\begin{itemize}
\item $\pi_C({\bf X}) = \pi_1 \circ \pi_2({\bf X})$,

\item $\pi_i({\bf X}) \supseteq \pi_C({\bf X})$ for $i \in [1..2]$.
\end{itemize}
Clearly, both properties hold when each function $f \in \C{\pi_C, \pi_1,
  \pi_2}$ is replaced by its canonic extension $f^+$ to the Cartesian
product $DO$ of all domains ${\cal P}(D_i)$. By the Stabilization
Corollary \ref{cor:chaotic}, Comparison Corollary \ref{cor:chaotic2}
and the counterpart of the {\tt CI} Theorem \ref{thm:CI} for the {\tt
  CIIQ} algorithm we conclude that both algorithms compute in $d$ the
same value.

An analogous analysis for arbitrary constraints allows us to compare
the algorithm of Davis \cite{davis87} discussed in
Subsection \ref{subsec:related} with that 
defined in Example \ref{exa:arccon}.
We can conclude that in case of termination both algorithms
achieve arc-consistency for $n$-ary constraints.

\subsection{Assessment and Future Work}

In this paper we showed that several constraint propagation
algorithms can be explained as simple instances of the chaotic iteration 
algorithms. Such a generic presentation also provides a framework for
generating new constraint propagation algorithms that can be 
tailored for specific application domains.
Correctness of these constraint propagation
algorithms does not have to be reproved each time anew.

It is unrealistic, however, to expect that all
constraint propagation algorithms presented in the literature can be
expressed as direct instances of the generic algorithms here considered.
The reason is that for some specific reduction functions
some additional properties of them can be exploited.

An example is the perhaps most known algorithm, the {\tt AC-3}
arc-consistency algorithm of Mackworth \cite{mackworth-consistency}.  We
found that its correctness relies in a subtle way on a commutativity
property of the projection functions discussed in Example
\ref{exa:arc}. This can be explained by means of a generic algorithm
only once one uses the information which function was applied last.

Another issue is that some algorithms, for example the {\tt AC-4}
algorithm of Mohr and Henderson \cite{MH86} and the {\tt GAC-4} algorithm of
Mohr and Masini \cite{MM88}, associate with each domain element some information
concerning its links with the elements of other domains. As a result
these algorithms operate on some ``enhancement'' of the original
domains.  To reason about these algorithms one has to relate the
original CSP to a CSP defined on the enhanced domains.

In an article under preparation we plan to discuss the
refinements of the general framework here presented that allow us
to prove correctness of such algorithms in a generic way.

\section*{Acknowledgements}
This work was prompted by our study of the first version of
van Emden \cite{Emd97}.  Rina Dechter helped us to clarify (most of) our
initial confusion about constraint propagation.  Discussions with Eric
Monfroy helped us to better articulate various points put forward
here. Nissim Francez, Dmitry Ushakov and both anonymous referees
provided us with helpful comments on previous versions of this paper.

\bibliographystyle{plain}

\bibliography{/ufs/apt/esprit/esprit,/ufs/apt/bib/clp2,/ufs/apt/esprit/chapter3,/ufs/apt/esprit/chapter4,/ufs/apt/book-lp/man1,/ufs/apt/book-lp/man2,/ufs/apt/book-lp/man3,/ufs/apt/book-lp/ref1,/ufs/apt/book-lp/ref2}

\end{document}

%% file: cmds.tex
\newcommand{\ES}{\mbox{$\emptyset$}}

\newcommand{\A}{\mbox{$\ \wedge\ $}}

\newcommand{\sse}{\mbox{$\:\subseteq\:$}}

\newcommand{\po}{\mbox{$\ \sqsubseteq\ $}}

\newcommand{\fa}{\mbox{$\forall$}}
\newcommand{\te}{\mbox{$\exists$}}

\newcommand{\LL}{\mbox{$\ldots$}}

\newcommand{\C}[1]{\mbox{$\{{#1}\}$}}           % curly braces

\newcommand{\NI}{\noindent}
\newcommand{\HB}{\hfill{$\Box$}}
\newcommand{\VV}{\vspace{5 mm}}
\newcommand{\III}{\vspace{3 mm}}
\newcommand{\II}{\vspace{2 mm}}

\newcommand{\ran}{\rangle}
\newcommand{\lan}{\langle}

\newcommand{\szkew}[1]{\relax \setbox0=\hbox{\kern -24pt $\displaystyle#1$\kern 0pt }%
\box0}
{\catcode`\@=11 \global\let\ifjusthvtest@=\iffalse}
\newcounter{oldmycaption}

\newcommand{\p}[2]{\langle #1 \ ; \ #2 \rangle}

\newcommand{\ceiling}[1]{\lceil #1 \rceil}
\newcommand{\floor}[1]{\lfloor #1 \rfloor}

%% file: headerlncs.tex
\def\smallromani{\renewcommand{\theenumi}{\roman{enumi}}
\renewcommand{\labelenumi}{(\theenumi)}}

\newcommand{\Proof}{\NI
		    {\bf Proof.}\ }
\newtheorem{claim}{Claim}

\newcounter{symbol}
\newcommand{\indexsyma}[1]%
{\stepcounter{symbol}\index{zzz1 \thesymbol @\protect#1}}
\newcommand{\indexsymb}[1]%
{\stepcounter{symbol}\index{zzz2 \thesymbol @\protect#1}}
\newcommand{\indexsymc}[1]%
{\stepcounter{symbol}\index{zzz3 \thesymbol @\protect#1}}
\newcommand{\indexsymd}[1]%
{\stepcounter{symbol}\index{zzz4 \thesymbol @\protect#1}}
\newcommand{\indexsyme}[1]%
{\stepcounter{symbol}\index{zzz5 \thesymbol @\protect#1}}